%% file: main.tex
\documentclass{article}
\ifdefined\pdfminorversion\pdfminorversion=7\fi
\usepackage{iclr2027_conference,times}
\usepackage[T1]{fontenc}
\usepackage[utf8]{inputenc}
\iclrfinaltrue
\usepackage{amsmath}
\usepackage{amssymb}
\usepackage{booktabs}
\usepackage{graphicx}
\usepackage{xcolor}
\usepackage[unicode]{hyperref}

\title{Hand2Robot: Object-Centric 4D\\Hand Reconstruction from Human\\Demonstrations for Robotics}
\input{authors}
\hypersetup{
    pdftitle={Hand2Robot: Object-Centric 4D Hand Reconstruction from Human Demonstrations for Robotics},
    pdfauthor={Zetao Du, Weiqing Wang, Hanbing Zhang, Yimiao Zhou, Ye Shi, Liang Wang, Jingya Wang}
}

\begin{document}
\maketitle

\input{sec/0_abstract}
\input{sec/1_intro_v1}
\input{sec/2_related_works}
\input{sec/3_method_unify}
\input{sec/4_experiment}
\input{sec/5_conclusion}
\input{sec/6_acknowledgments}

\bibliographystyle{iclr2027_conference}
\bibliography{example}
\clearpage
\input{sec/7_appendix}
\end{document}

%% file: authors.tex
\author{
\textbf{Zetao Du\textsuperscript{1}\thanks{Equal contribution.}
\hspace{1.5em}
Weiqing Wang\textsuperscript{1,2}\footnotemark[1]
\hspace{1.5em}
Hanbing Zhang\textsuperscript{1}
\hspace{1.5em}
Yimiao Zhou\textsuperscript{1}} \\
\textbf{Ye Shi\textsuperscript{1,2}
\hspace{1.5em}
Liang Wang\textsuperscript{3}
\hspace{1.5em}
Jingya Wang\textsuperscript{1}\thanks{Corresponding author.}} \\[0.3em]
\textsuperscript{1}ShanghaiTech University
\qquad
\textsuperscript{2}InstAdapt \\
\textsuperscript{3}Institute of Automation, Chinese Academy of Sciences \\[0.3em]
{\small\texttt{\{duzt2022,zhanghb2025,zhouym2025,shiye,wangjingya\}@shanghaitech.edu.cn}} \\
{\small\texttt{wangweiqing@instadapt.com}\quad\texttt{wangliang@nlpr.ia.ac.cn}}
}

%% file: sec/0_abstract.tex

\begin{abstract}
Human demonstrations can capture task-relevant manipulation without using the target robot, but
the recorded hand motions cannot directly serve as robot action supervision.
Hand reconstruction and motion retargeting provide a direct route to such supervision, yet
recovering hand motion alone does not ensure physically consistent hand-object interactions.
To address this issue, we introduce Hand2Robot, an object-centric 4D hand reconstruction framework for robot policy post-training. Given videos of humans performing the target task, Hand2Robot extracts interaction clips, reconstructs object geometry, and generates robot action supervision through hand-object interaction optimization and motion retargeting.
For object tracking, we combine observation confidence with
temporal motion constraints to select a globally consistent object pose trajectory from per-frame
pose candidates. For hand refinement, we use object geometry and motion as a reference and combine
staged 2D alignment with object-local contact constraints to refine hand trajectories and finger
joint configurations. Experiments on DexYCB and HO3D-v3 show
improved object pose accuracy and hand-object interaction quality. 
Experiments on three real-robot dexterous manipulation tasks further
show that reconstructed and retargeted human demonstrations can
effectively supplement limited robot data for target-task policy
post-training.
Project~page: \url{https://Hand2Robot.github.io/}.
\end{abstract}

%% file: sec/1_intro_v1.tex

\section{Introduction}
\label{sec:introduction}

Deploying pretrained robot policies for dexterous manipulation on a target robot
often requires post-training with task-relevant demonstrations.
Collecting robot demonstrations at low cost remains challenging, given constraints
such as dependence on the target hardware and the need for skilled teleoperators.
These constraints motivate the use of human manipulation
videos~\citep{grauman2022ego4d,hoque2025egodex} as a complementary data source:
humans can directly demonstrate task-relevant object interactions without
operating the target robot. Using these videos for policy post-training requires
translating human motions into the robot's action space while preserving the
demonstrated interactions. We therefore investigate how to generate reliable
robot-action supervision from human demonstrations to complement limited
robot data.

Existing research uses human videos to support robot learning through
visual representations, reward signals, or latent action
pretraining~\citep{hu2024video,guzey2025hudor,ye2025latent}.
Another line of work recovers human motion from video and generates
robot-action supervision through motion alignment or
retargeting~\citep{kareer2025egomimic,yang2025egovla}.
For dexterous manipulation, recovering hand motion also requires
preserving the demonstrated hand-object relationships: object geometry
constrains feasible hand positions and joint configurations relative
to the object, while contact relationships describe how the hand
interacts with it~\citep{fan2024hold,xu2026choir}.
Ignoring these relationships can produce visually plausible motion
with missing contacts or hand-object interpenetration, compromising
the reliability of the generated supervision.
Prior work improves manipulation trajectories recovered from video
through geometric refinement, contact optimization, or interaction-state
constraints~\citep{chen2026dexterous,egoinfinity_2026,Zhang_2026_CVPR}.



However, refining hand motion under geometric and contact constraints
still requires reliable estimates of object geometry and pose.
Occlusion and pose ambiguity can introduce errors and temporal
inconsistencies into object pose
estimates~\citep{paliwal2026dexterous}, potentially leading hand
refinement away from the demonstrated interaction.
Moreover, locally plausible contact does not ensure that hand motion
agrees with image observations and remains temporally coherent.
To address these challenges, we introduce Hand2Robot, an object-centric
4D hand reconstruction framework that combines global object trajectory
selection with hand refinement constrained by observations and contact.
Given RGB-D videos of humans performing the target task, Hand2Robot
reconstructs object geometry and combines observational evidence with
temporal continuity to select a globally consistent object pose
trajectory from per-frame pose candidates.
Using the recovered object geometry and motion as references, it then
jointly refines the hand's global pose trajectory and finger joint
configurations using image alignment together with contact and temporal
constraints.
Finally, the refined motion is retargeted to the target robot's action
space to generate action supervision for policy post-training.

\begin{figure}[!t]
    \centering
    \includegraphics[width=\linewidth]{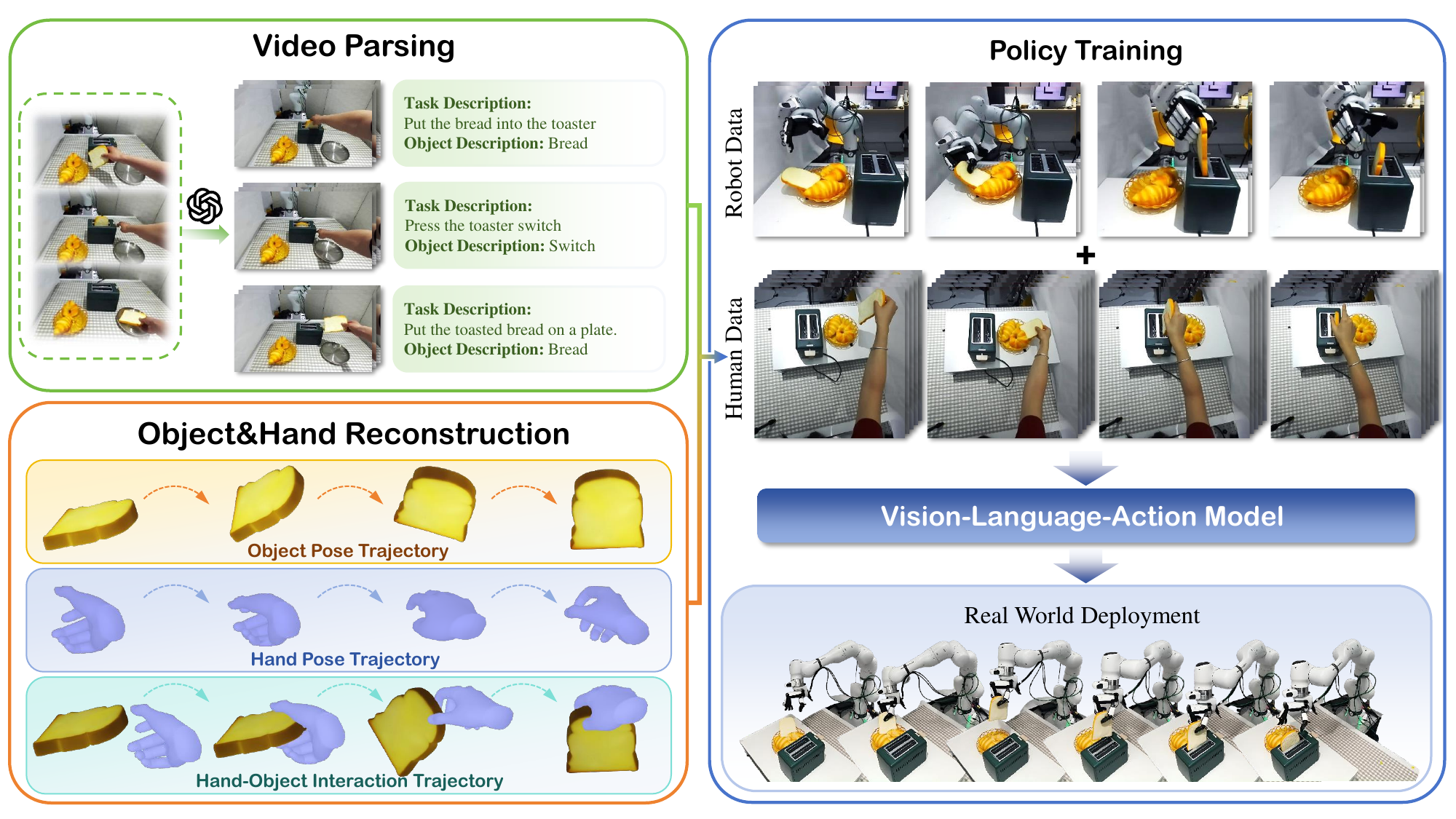}
    \caption{Overview of the Hand2Robot pipeline. Given human demonstration videos, we first parse them
    into several semantic tasks by VLM. Then we apply our hand-object reconstruction and refinement
    framework for physically realistic hand trajectories. Finally, these hand trajectories are
    retargeted to the robot action space, enabling human-robot data co-training on a
    Vision-Language-Action model for real-world deployment.}
    \label{fig:pipeline}
\end{figure}

During this reconstruction process, object pose is affected by hand occlusion and object symmetry,
while hand motion is affected by partial occlusion and contact ambiguity~\citep{fan2024hold,xu2026choir}.
For object tracking, selecting a single pose independently in each frame can introduce errors under
occlusion and accumulate drift, or cause frequent switching between equivalent poses of symmetric
objects. Hand2Robot therefore retains multiple pose candidates per frame and combines their
observation confidence with temporal continuity between adjacent frames to select a globally
consistent object pose trajectory.
For hand reconstruction, generic hand estimators often produce incomplete estimates in contact
regions and do not use object geometry, so their outputs serve only as initial
trajectories~\citep{fan2024hold,xu2026choir}. Hand2Robot uses contact-aware staged optimization:
it first stabilizes hand placement through 2D observation alignment, then introduces contact
consistency constraints in the object coordinate frame and penetration penalties to suppress
hand-object interpenetration and abrupt changes in contact location.

We evaluate Hand2Robot in terms of reconstruction quality and downstream policy learning. First, on
public hand-object interaction benchmarks including DexYCB~\citep{chao2021dexycb} and
HO3D-v3~\citep{hampali2020honnotate}, we evaluate object pose, object geometry, hand motion,
hand-object interaction quality, and temporal stability under a common evaluation protocol,
comparing against initialization results and baseline methods. Second, on three real-robot dexterous
manipulation tasks, we combine reconstructed and retargeted human demonstrations with limited robot
demonstrations for policy post-training, and compare against training with robot demonstrations
alone to examine whether the generated human data provides useful action supervision.

Our main contributions are as follows:
\begin{itemize}
\item We introduce Hand2Robot, a framework that converts human demonstration videos into
action supervision for dexterous robot policy post-training. The framework combines
hand-object interaction reconstruction with motion retargeting, enabling human
demonstrations to complement limited robot data.
\item We propose an object-centric 4D hand reconstruction pipeline that
estimates object geometry and pose trajectories, then uses them as fixed
references to refine hand motion. By combining confidence-aware temporal object
tracking with observation-to-contact hand refinement, our method maintains interaction consistency
under occlusion, object symmetry, and contact ambiguity.
\item We demonstrate the effectiveness of the generated action supervision
for policy post-training on three real-robot dexterous manipulation tasks.
The results show that reconstructed and retargeted human demonstrations
can effectively complement limited robot data.
\end{itemize}

%% file: sec/2_related_works.tex

\section{Related Works}
\label{sec:related_works}

\paragraph{Robot Learning from Human Videos.}
Human videos offer a scalable complement to robot teleoperation data, but their utility for policy
learning depends on whether visual demonstrations can be grounded into action-like supervision.
Prior works exploit human videos for representation
learning~\citep{ma2022vip,nair2022r3m,radosavovic2023real}, trajectory or plan
extraction~\citep{bharadhwaj2024track2act,wen2023any,wang2023mimicplay}, cross-embodiment
retargeting~\citep{qin2022dexmv,chen2022dextransfer,chen2025vividex}, and action-space learning for
vision-language-action models~\citep{ye2025latent,bu2025univla,bjorck2025gr00t}. More recent
egocentric approaches recover hand motion or interaction cues to bridge human demonstrations and
robot manipulation~\citep{mandikal2022dexvip,yang2025egovla,luo2025being,bi2026h}. However, raw
videos usually lack physically grounded hand-object trajectories, making it difficult to mix them
directly with robot demonstrations. Our work addresses this gap by reconstructing HOI trajectories
from human videos and using the resulting annotations as additional training data for improving
policies such as $\pi_{0.5}$~\citep{intelligence2025pi_}.

\paragraph{Hand-Object Pose Estimation.}
Recent hand foundation models such as HaMeR and WiLoR provide strong monocular hand mesh
initialization~\citep{pavlakos2024reconstructing,potamias2025wilor}, while video methods improve
temporal hand trajectories~\citep{yu2025dyn,ye2025predicting,zhang2025hawor}. Yet hand-only
estimates lack object geometry and contact constraints, causing depth drift, penetration, and
unstable contact under occlusion. Hand-object reconstruction methods further introduce object
information to constrain the interaction. Template-based or RGB-D/multi-view methods use explicit
object geometry and
pose~\citep{tekin2019h+,liu2021semi,grady2021contactopt,yang2021cpf,chao2021dexycb,hampali2020honnotate,yang2022oakink},
while recent template-free methods use implicit fields, SDFs, NeRFs, diffusion models, or foundation
models to recover unknown objects and hand-object
configurations~\citep{hasson2019learning,karunratanakul2020grasping,chen2022alignsdf,chen2023gsdf,fan2024hold,ye2023diffusion,wang2025magichoi,liu2025easyhoi}.
These works show that object geometry and contact constraints are essential, but reliable
hand-object trajectories remain challenging in human manipulation videos due to occlusion, depth ambiguity,
and unstable initialization. Building on these advances, we use foundation models to initialize
object and hand states, and introduce two refinement modules: confidence-guided object pose
trajectory estimation for the object, and observation-to-contact hand refinement for the hand.

%% file: sec/3_method_unify.tex

\section{Method}
\label{sec:method}

\begin{figure}[!t]
    \centering
    \includegraphics[width=\linewidth]{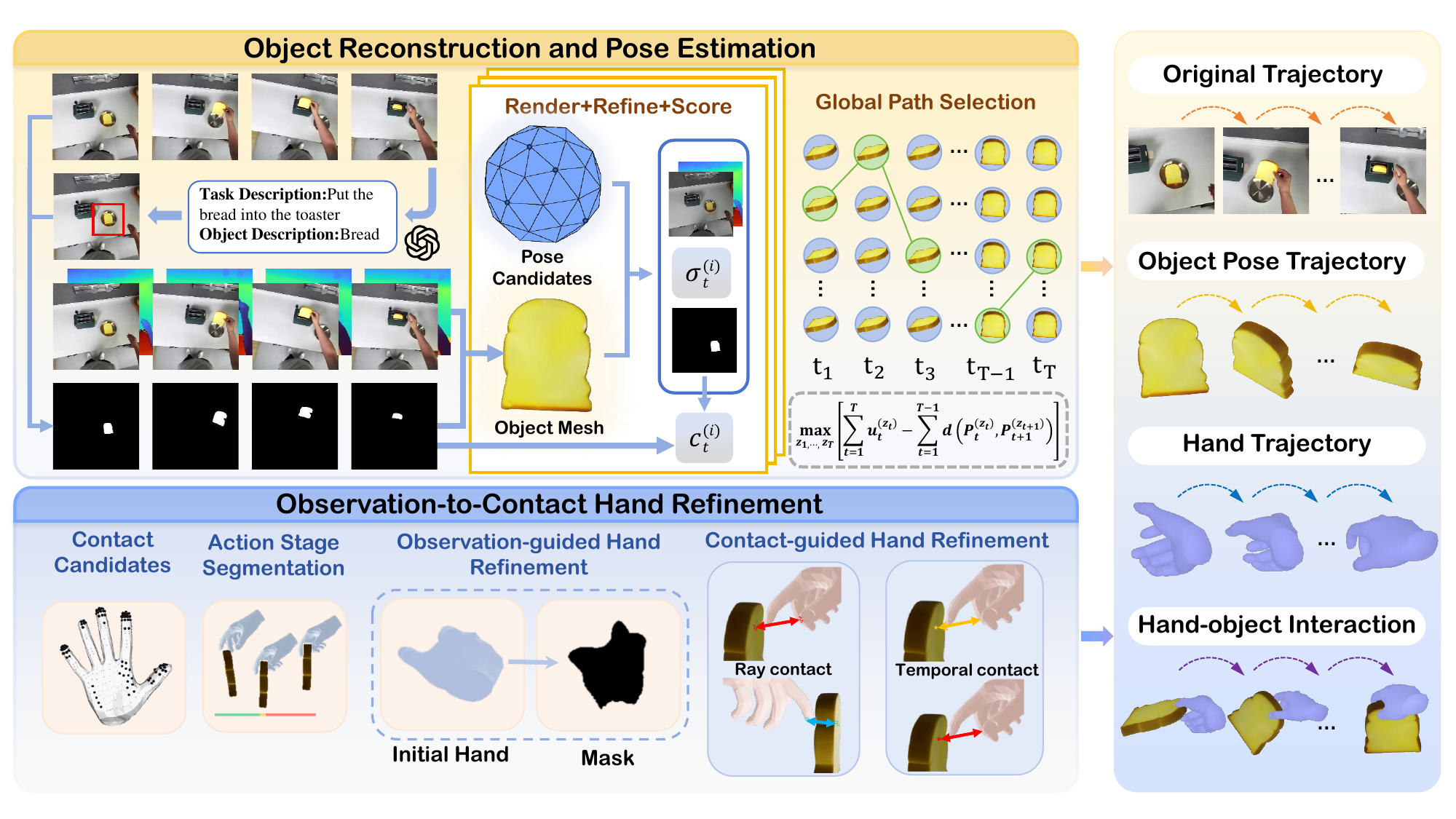}
\caption{Overview of our object-centric 4D hand reconstruction pipeline.}
    \label{fig:hoi}
\end{figure}


We consider the problem of converting human demonstrations into action supervision for dexterous
robot policy post-training. The input consists of RGB-D videos of humans performing the target
manipulation task, together with the corresponding camera intrinsics. Each video may contain one
or more interaction episodes. Hand2Robot reconstructs hand-object interactions and retargets the
refined hand motion to the target robot, producing a sequence of robot-action labels temporally
aligned with the video observations. The resulting observation-action pairs serve as supplementary
training data alongside limited robot demonstrations.

\subsection{Video Parsing and Object Grounding}
\label{subsec:data_processing}
We first use a vision-language model (VLM)~\citep{singh2025openai} to segment the input videos
into interaction clips and annotate each clip with descriptions of the task and manipulated object. 
For each clip, we use the object description to guide Rex-Omni~\citep{jiang2025detect}
in localizing the manipulated object. The resulting detections serve as prompts for
SAM3~\citep{carion2025sam}, yielding per-frame object masks $\{M_t\}_{t=1}^T$
for subsequent object reconstruction and observation-based scoring of pose candidates.
We also use MapAnything~\citep{keetha2025mapanything} to
recover camera extrinsics for each RGB-D frame.

\subsection{Object Reconstruction and Robust Pose Tracking}
\textbf{Object Reconstruction.} We use SAM~3D~\citep{chen2025sam} to reconstruct
the manipulated object as a metric-scale 3D mesh $\mathcal{O}$ from an RGB image and its
corresponding object mask. The reconstructed mesh provides a geometric reference
for subsequent object pose tracking and hand-object interaction refinement.

\textbf{Sequential Candidate Selection.}
Hand occlusion and object symmetry can make per-frame object pose estimates ambiguous.
We therefore retain the $N$ refined pose candidates from
FoundationPose~\citep{wen2024foundationpose} at each frame, together with their ScoreNet scores.
This yields the candidate set $\mathcal{P}_t = \{P_t^{(i)}\}_{i=1}^N$, where
$P_t^{(i)} = (\mathbf{R}_t^{(i)},\mathbf{t}_t^{(i)})$ denotes the $i$-th pose candidate
at frame $t$, with raw score $\sigma_t^{(i)}$. We then select an object pose trajectory
across the full sequence by combining observation agreement with temporal continuity.

We associate each candidate $P_t^{(i)}$ with a unary weight $u_t^{(i)} \in [0,1]$ that measures its
agreement with the visual observation, and define the transition cost between adjacent-frame
candidates as:
\begin{equation}
d(P_t^{(i)}, P_{t+1}^{(j)}) = \lambda_{\mathrm{trans}} \|\mathbf{t}_t^{(i)} -
\mathbf{t}_{t+1}^{(j)}\|_2 + \lambda_{\mathrm{rot}}
\arccos\!\left(\frac{\operatorname{tr}\big((\mathbf{R}_t^{(i)})^\top
\mathbf{R}_{t+1}^{(j)}\big)-1}{2}\right),
\end{equation}
where $\lambda_{\mathrm{trans}},\lambda_{\mathrm{rot}}>0$ control the translation and rotation
smoothness, respectively. The full optimization reads:
\begin{equation}
\max_{z_1,\ldots,z_T} \left[ \sum_{t=1}^{T} u_t^{(z_t)} - \sum_{t=1}^{T-1} d(P_t^{(z_t)},
P_{t+1}^{(z_{t+1})}) \right].
\label{eq:seq-opt}
\end{equation}
The optimization in Eq.~(\ref{eq:seq-opt}) selects one candidate index $z_t \in \{1,\dots,N\}$ per
frame across all $T$ frames. This induces a graph with $T$ layers and $N$ nodes per layer: node $i$
at layer $t$ carries the unary weight $u_t^{(i)}$, and a directed edge from node $i$ at layer $t$ to
node $j$ at layer $t+1$ carries weight $-d\bigl(P_t^{(i)}, P_{t+1}^{(j)}\bigr)$. Since all edges
point strictly from $t$ to $t+1$, the graph is acyclic and layered, reducing Eq.~(\ref{eq:seq-opt})
to a longest-path problem solvable by Viterbi decoding~\citep{rabiner1989tutorial} in $O(T N^2)$.

\textbf{Construction of Unary Weights.}
We use the ScoreNet scores $\{\sigma_t^{(i)}\}$ to rank pose candidates within each frame and
apply per-frame min-max normalization:
\begin{equation}
\bar{\sigma}_t^{(i)} = \frac{\sigma_t^{(i)} - \min_j \sigma_t^{(j)}}{\max_j \sigma_t^{(j)} - \min_j
\sigma_t^{(j)}},
\end{equation}
This normalization preserves the within-frame ranking, but the highest-scoring candidate receives
a value of 1 even when all candidates are inaccurate. We therefore supplement the normalized
scores with image-space agreement. For each candidate $P_t^{(i)}$, we render the object mask and
compute its IoU with the observed mask $M_t$:
\begin{equation}
c_t^{(i)} = \operatorname{IoU}\big(\mathcal{R}_{\mathrm{mask}}(\mathcal{O}, P_t^{(i)}), M_t\big),
\end{equation}
The unary weight combines the normalized ScoreNet score with this observation-agreement weight:
\begin{equation}
u_t^{(i)} = c_t^{(i)} \cdot \bar{\sigma}_t^{(i)}.
\end{equation}
This weighting reduces the contribution of candidates whose rendered masks disagree with the
observation. When mask agreement is low for all candidates in a frame, the reduced unary weights
increase the relative influence of temporal transition costs in Eq.~(\ref{eq:seq-opt}) on
trajectory selection.

\subsection{Observation-to-Contact Hand Refinement}

We obtain per-frame MANO estimates~\citep{MANO:SIGGRAPHASIA:2017} from
WiLoR~\citep{potamias2025wilor} and use tracking and depth-based alignment for metric
initialization. Using the reconstructed object geometry and pose trajectory as fixed references,
we jointly refine the hand's global pose trajectory and finger joint configurations across
the sequence. The refinement emphasizes 2D observation alignment before reliable contact
is available and incorporates contact constraints during interaction.

\textbf{Action stage segmentation.}
We segment each action into grasping and interaction stages using object motion and hand--object
proximity. Let $T_t^o$ be the object translation and $d_t^{\mathrm{ho}}$ the minimum hand--object
distance:
\begin{equation}
e_t=
\alpha\,\frac{\|T^o_t-T^o_{t-1}\|_2}{\delta_{\mathrm{trans}}}
+\beta\,\max\left(0,1-\frac{d_t^{\mathrm{ho}}}{\delta_{\mathrm{ho}}}\right),
\end{equation}
where $\alpha,\beta$ are cue weights and $\delta_{\mathrm{trans}},\delta_{\mathrm{ho}}$ are
normalization thresholds. We select the earliest frame satisfying $e_t>\delta_{\mathrm{on}}$ as the
anchor $t_a$. Single-frame refinement at $t_a$ yields a hand translation offset
$\Delta\tau=\tau_a^\star-\tau_a^0$, where $\tau_a^0$ is the initial hand translation at the anchor
frame and $\tau_a^\star$ is the refined translation after single-frame optimization. This offset
initializes the entire segment.

We then optimize the MANO parameters $\Theta_t=\{s_t,\tau_t,\omega_t,\theta_t,\beta_t\}$
(see Appendix~\ref{app:hand_refinement}) while keeping the object pose
$\mathcal{O}_t=(R_t^o,T_t^o)$ fixed:
\begin{equation}
\mathcal{L}=
\lambda_{\mathrm{2d}}\mathcal{L}_{\mathrm{2d}}
+\lambda_{\mathrm{match}}\mathcal{L}_{\mathrm{match}}
+\lambda_{\mathrm{temp}}\mathcal{L}_{\mathrm{temp}}
+\lambda_{\mathrm{sdf}}\mathcal{L}_{\mathrm{sdf}}
+\lambda_{\mathrm{sm}}\mathcal{L}_{\mathrm{sm}}
+\lambda_{\mathrm{reg}}\mathcal{L}_{\mathrm{reg}} .
\end{equation}
Here, $\mathcal{L}_{\mathrm{2d}}$ aligns masks,
$\mathcal{L}_{\mathrm{match}}$ encourages per-frame contact matching,
$\mathcal{L}_{\mathrm{temp}}$ promotes temporal contact consistency,
$\mathcal{L}_{\mathrm{sdf}}$ penalizes penetration, and
$\mathcal{L}_{\mathrm{sm}}+\mathcal{L}_{\mathrm{reg}}$ regularize the trajectory.
All $\lambda_{(\cdot)}$ are scalar weights.

\textbf{2D-guided hand alignment.}
During the grasping stage, we emphasize 2D alignment before reliable contact constraints
are available. We use the stage-weighted mask loss
$\mathcal{L}_{\mathrm{2d}}=\frac{1}{B}\sum_t
\gamma_t\,\operatorname{SoftIoU}(\hat{M}_h^t,M_h^t)$,
where $B$ is the number of optimized frames, $\hat{M}_h^t$ and $M_h^t$ are the rendered
and observed hand masks, and $\operatorname{SoftIoU}$ is a differentiable soft IoU loss.
We set $\gamma_t=\gamma_g>1$ for grasping frames and $\gamma_t=1$ for interaction frames.

\textbf{Per-frame contact matching.}
During interaction, we adopt ray-based 2D-to-3D contact matching~\citep{lin2025zero}.
For each active MANO contact candidate $v_{t,k}$, a camera ray through its image
projection intersects the posed object mesh at $o_{t,k}$.
We minimize the matching loss
\begin{equation}
\mathcal{L}_{\mathrm{match}}=
\frac{1}{|\mathcal{I}_{\mathrm{inter}}|}
\sum_{t\in\mathcal{I}_{\mathrm{inter}}}
\frac{1}{\max(|\mathcal{C}_t|,1)}
\sum_{k\in\mathcal{C}_t}\|v_{t,k}-o_{t,k}\|_2 ,
\end{equation}
where $\mathcal{I}_{\mathrm{inter}}$ is the interaction-frame set and $\mathcal{C}_t$ is the active
contact set.

\textbf{Temporal contact stabilization.}
Per-frame matching can assign the same hand candidate to different object regions
in adjacent frames. To improve temporal consistency, we regularize contact
locations in the object coordinate frame, accounting for the object's rigid motion.
We transform the ray-hit point and hand
candidate as $\bar{o}_{t,k}=(R_t^o)^\top(o_{t,k}-T_t^o)$ and
$\bar{v}_{t,k}=(R_t^o)^\top(v_{t,k}-T_t^o)$. For candidates active in two adjacent frames,
$\mathcal{C}_{t,t+1}=\mathcal{C}_t\cap\mathcal{C}_{t+1}$, we use the average ray-hit location as a
stop-gradient local anchor,
$a_{t,k}=\frac{1}{2}(\operatorname{sg}(\bar{o}_{t,k})+\operatorname{sg}(\bar{o}_{t+1,k}))$, and
minimize
\begin{equation}
\mathcal{L}_{\mathrm{temp}} =
\sum_{t,\,t+1\in\mathcal{I}_{\mathrm{inter}}}
\sum_{k\in\mathcal{C}_{t,t+1}}
\left[
\rho(\|\bar{v}_{t,k}-a_{t,k}\|_2)
+\rho(\|\bar{v}_{t+1,k}-a_{t,k}\|_2)
\right].
\end{equation}
Here $\operatorname{sg}$ is stop-gradient and $\rho(x)=\sqrt{x^2+\epsilon}$ is the Charbonnier
robust loss with a small $\epsilon>0$ for numerical stability. 
This term encourages the same hand candidate to remain near a common local
contact region across adjacent frames, reducing inconsistent contact drift.

After warmup, we enable $\mathcal{L}_{\mathrm{sdf}}$ to penalize hand--object
penetration, while $\mathcal{L}_{\mathrm{sm}}$ suppresses abrupt motion changes
and $\mathcal{L}_{\mathrm{reg}}$ limits deviations from the initialization.
Further details are provided in Appendix~\ref{app:hand_refinement}.

\subsection{From Human Hand to Dexterous Hand}
The final step converts the refined MANO motion into robot action supervision.
We use GeoRT~\citep{yin2025geometric} to obtain joint configurations for the
target dexterous hand. The global hand pose trajectory is first transformed
from camera to world coordinates using the camera extrinsics from
Sec.~\ref{subsec:data_processing}. We then use representative initial object
poses in the human and robot demonstrations to estimate a world-to-robot
transformation and map the trajectory into the robot frame. The aligned poses and retargeted joint
configurations are encoded using the same action representation as robot
demonstrations and paired with the corresponding video observations for
policy post-training.

%% file: sec/4_experiment.tex

\section{Experimental Results}
\label{sec:experiment}

Our experiments evaluate Hand2Robot along two axes. First, we evaluate whether the proposed
object-centric reconstruction improves hand-object trajectory quality on public benchmarks. Second,
we evaluate whether the reconstructed and retargeted human demonstrations improve policy learning in
real-robot dexterous manipulation tasks.

\subsection{Hand-Object Reconstruction Evaluation}

We first evaluate the reconstruction component of Hand2Robot on two public hand-object interaction
benchmarks. We assess object reconstruction quality, the accuracy of global object pose-path
selection, and the quality of hand-object trajectories after object-centric hand refinement.

\textbf{Datasets and Metrics.} We evaluate on DexYCB~\citep{chao2021dexycb} and
HO3D-v3~\citep{hampali2020honnotate}. For DexYCB, we select a subset of 1,000 sequences for object
pose evaluation, from which we further sample 120 sequences for hand evaluation.
HO3D-v3 contains longer manipulation sequences. We sample a subset of 12 sequences
for evaluation. For object pose, we report translation error, rotation error, and ADD/ADD-S
AUC~\citep{xiang2017posecnn}. We evaluate hand pose accuracy (e.g., MPJPE, MRPE), physical
interaction metrics (e.g., penetration volume, IV), and jerk magnitude (JM) to compare Hand2Robot
against the WiLoR baseline.
Metric definitions and alignment procedures are described in
Appendix~\ref{app:evaluation_protocal}.

\textbf{Object Reconstruction Quality.} We additionally evaluate object reconstruction on DexYCB.
Table~\ref{tab:object_fidelity} in Appendix~\ref{app:object_reconstruction_quality} reports CD,
F@5, and F@10 for HOLD, MagicHOI, and our pipeline. Our pipeline achieves a CD of 0.57~cm$^2$,
with F@5 and F@10 of 49.35\% and 87.54\%, respectively.

\textbf{Object Pose Tracking Results.} Table~\ref{tab:object} reports object pose tracking results
on DexYCB. Compared with the default FoundationPose output, our global pose-path selection reduces
translation error from 6.80 to 3.44~cm and rotation error from 27.14$^\circ$ to 18.47$^\circ$.
ADD AUC increases from 98.96 to 99.48, while ADD-S AUC increases from 99.15 to 99.62.
These results show improved object pose accuracy when selecting the trajectory using both
observation scores and temporal constraints.

\input{table/object}

\input{sec/opt_exp}

\textbf{Hand Pose and Trajectory Refinement Results.} Table~\ref{tab:comparison} reports results on
DexYCB~\citep{chao2021dexycb} and HO3D-v3~\citep{hampali2020honnotate}. While WiLoR provides strong
per-frame initialization, it lacks temporal and contact constraints, leading to trajectory drift and
object penetration. By integrating 2D mask alignment, ray-based contact, SDF, and temporal
smoothness, our optimization yields temporally consistent and physically plausible trajectories.
Specifically, our refinement reduces wrist trajectory error (MRPE)
on DexYCB and hand-object interpenetration (IV) on both datasets.
These results show reduced hand-object interpenetration and improved
wrist trajectory accuracy and smoothness.

\input{sec/4_robot_experiments}

%% file: table/object.tex

\begin{table}[!htbp]
\caption{Quantitative object pose results on DexYCB using reconstructed meshes and first-frame
alignment. See Appendix~\ref{app:evaluation_protocal} for evaluation details.}
\label{tab:object}
\centering
\resizebox{\linewidth}{!}{%
\begin{tabular}{ccccc}
\toprule
Method & Translation error (cm) $\downarrow$ & Rotation error ($^\circ$) $\downarrow$ & ADD AUC $\uparrow$ & ADD-S AUC $\uparrow$ \\
\midrule
FoundationPose~\citep{wen2024foundationpose} & 6.80 & 27.14 & 98.96 & 99.15 \\
\midrule
Ours w/o $d,c$ & 4.88 & 61.63 & 99.07 & 99.44 \\
Ours w/o $d$   & 3.62 & 38.20 & 99.37 & 99.61 \\
Ours w/o $c$   & \textbf{3.44} & 19.86 & 99.47 & \textbf{99.62} \\
Ours           & \textbf{3.44} & \textbf{18.47} & \textbf{99.48} & \textbf{99.62}\\
\bottomrule
\end{tabular}
}
\end{table}

%% file: sec/opt_exp.tex

\begin{table}[!htbp]
  \centering
  \small
  \setlength{\tabcolsep}{5pt}
\caption{Quantitative evaluation for in-scene object manipulation. We evaluate on HO3D-v3 and DexYCB,
comparing the WiLoR initialization with our refined hand trajectories. We highlight the best values
in bold.}
  \label{tab:comparison}
  \resizebox{\linewidth}{!}{%
\begin{tabular}{@{}llccccc@{}}
    \toprule
    \textbf{Dataset} & \textbf{Method}
      & MPJPE $\downarrow$
      & MRPE $\downarrow$
      & IV-mean $\downarrow$
      & IV-max $\downarrow$
      & JM $\downarrow$ \\
    \midrule
    \textbf{DexYCB}~\citep{chao2021dexycb}
      & WiLoR~\citep{potamias2025wilor}                         & 8.35 & 97.31 & 2.57 & 5.41 & 2.58 \\
      & Ours                           & \textbf{7.58} & \textbf{38.49} & \textbf{1.07} & \textbf{3.25} & \textbf{2.49} \\
    \midrule
    \textbf{HO3D-v3}~\citep{hampali2020honnotate}
      & WiLoR                          & \textbf{4.22} & 24.22 & 20.11 & 64.82 & 0.86 \\
      & Ours                           & 4.40 & \textbf{16.99} & \textbf{3.42} & \textbf{22.40} & \textbf{0.34} \\
    \bottomrule
  \end{tabular}
}
\end{table}

%% file: sec/4_robot_experiments.tex

\subsection{Real-Robot Experiments}
\label{sec:real_robot}

\subsubsection{Experimental Setup and Evaluation Protocol}

We evaluate whether the action supervision generated by Hand2Robot supports policy post-training for
target tasks on a real robot. Our platform consists of a Franka arm and an xHand dexterous hand.
Human and robot demonstrations are recorded using the same fixed side-view camera system. Human
RGB-D demonstrations are converted into action supervision in the target robot coordinate frame
through hand-object interaction reconstruction and motion retargeting, and are combined with limited
robot demonstrations for $\pi_{0.5}$ post-training. This setup allows us to examine how the
processing of human demonstrations affects downstream policy performance under a controlled
observation viewpoint.

\begin{figure}[!htbp]
    \centering
    \includegraphics[width=\linewidth]{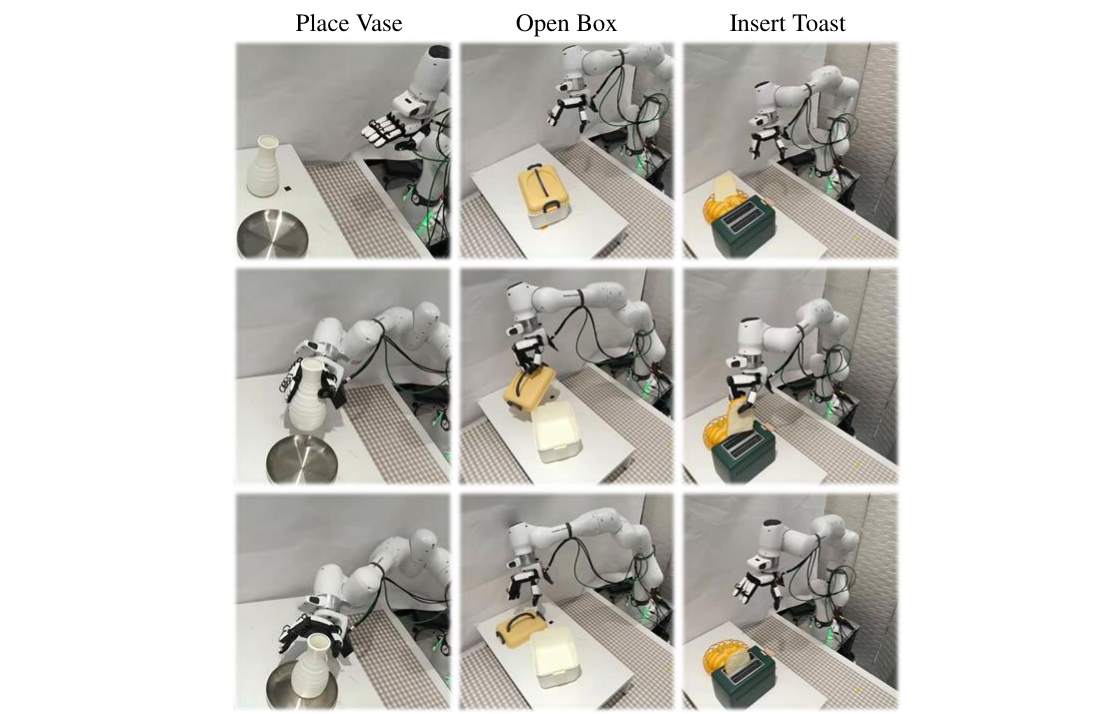}
    \caption{Examples of real-robot manipulation tasks. Each column shows one task, with the three
    rows corresponding to its initial, intermediate, and successful final states. The tasks are
    Place Vase, Open Box, and Insert Toast, from left to right.}
    \label{fig:task}
\end{figure}

Our main comparisons cover three real-robot tasks shown in Figure~\ref{fig:task}: Place Vase, Open Box, and Insert Toast, which
assess policy performance in pick-and-place, opening, and insertion, respectively. All conditions
use the same policy architecture and training procedure. In the main comparison of action
generation methods, we hold the 20 robot demonstrations and 100 source human videos per task
fixed, while varying how human action supervision is generated. The separate budget study varies
the number of demonstrations. During mixed-data post-training, human and robot training samples
are drawn at a 1:1 ratio. Further details are provided in
Appendix~\ref{app:real_world_robot_experiment_details}.

We use task success rate as the primary evaluation metric. Each trained policy is evaluated in
20 trials per task, with initial object positions randomly sampled from the training distribution.
Success rate is the number of successful trials divided by 20. We report success rates separately
for Place Vase, Open Box, and Insert Toast, and use their unweighted mean to summarize overall
performance.

\subsubsection{Downstream Effects of Human Action Supervision}

We compare WiLoR with basic retargeting, MagicHOI, and Hand2Robot using the same
100 human demonstrations and 20 robot demonstrations per task for $\pi_{0.5}$
post-training. MagicHOI uses its hand optimization module within a common pipeline;
robot-only training serves as a reference.

\input{table/robot_control}

Table~\ref{tab:robot} shows a mean success rate of 80.0\% for Hand2Robot,
compared with 18.3\% for WiLoR, 26.7\% for MagicHOI, and 48.3\% for robot-only
training. Hand2Robot outperforms both human-supervision baselines on all three tasks
and improves over robot-only training by 35, 25, and 35 percentage points on
Place Vase, Open Box, and Insert Toast, respectively. These results show that
Hand2Robot generates more effective action supervision than the compared methods
and improves policy post-training with limited robot data.

\subsubsection{Robot Demonstration Budget and the Value of Human Data}

We further compare 20R+100H with robot-only budgets of 20R, 40R, and 60R to assess
whether Hand2Robot-generated supervision can reduce robot demonstration
requirements for policy post-training.

\input{table/robot_budget}

As shown in Table~\ref{tab:robot_budget}, training with 20R+100H achieves a mean
success rate of 80.0\%, approaching the 81.7\% obtained with 60R and exceeding
the 48.3\% and 60.0\% obtained with 20R and 40R, respectively.
At the task level, 20R+100H matches the success rates of 60R on Place Vase and
Insert Toast, and is five percentage points lower on Open Box.
These results demonstrate the value of Hand2Robot-generated supervision for
achieving high task success rates with a smaller robot demonstration budget.

We further examine the effect of human demonstration quantity while fixing the
robot demonstration budget at 20. Increasing the number of human demonstrations
from 0 to 50 and 100 raises the mean success rate from 48.3\% to 55.0\% and
80.0\%, respectively. Within the evaluated range, additional human demonstrations
processed by Hand2Robot provide further benefits to policy post-training.
We additionally evaluate Do As I Do~\citep{paliwal2026dexterous}, retaining its
dynamics-based post-optimization and adapting its Sharpa outputs to xHand.
At 20R+50H, this adapted baseline achieves success rates of 50.0\%, 45.0\%,
and 40.0\% on Place Vase, Open Box, and Insert Toast, respectively, with a mean
of 45.0\%, compared with 55.0\% for Hand2Robot under the same demonstration budget.

Training with 100H alone achieves a mean success rate of 8.3\%, compared with
80.0\% for 20R+100H, highlighting the benefit of combining human and robot
demonstrations in this setting.

%% file: table/robot_control.tex

\begin{table}[htbp]
\centering
\small
\setlength{\tabcolsep}{4pt}
\caption{Task success rates (\%) after policy post-training with different sources of human action
supervision. R and H denote the numbers of robot and human demonstrations per task.
The mixed-data conditions use the same 20 robot demonstrations and 100 source human videos.
Mean is the unweighted average across the three tasks.}
\label{tab:robot}
\begin{tabular}{@{}lrrrr@{}}
\toprule
Training data and method & Place Vase & Open Box & Insert Toast & Mean \\
\midrule
20R & 45.0 & 60.0 & 40.0 & 48.3 \\
20R + 100H, WiLoR & 25.0 & 20.0 & 10.0 & 18.3 \\
20R + 100H, MagicHOI & 40.0 & 25.0 & 15.0 & 26.7 \\
20R + 100H, Hand2Robot & 80.0 & 85.0 & 75.0 & 80.0 \\
\bottomrule
\end{tabular}
\end{table}

%% file: table/robot_budget.tex

\begin{table}[htbp]
\centering
\small
\setlength{\tabcolsep}{4pt}
\caption{Success rates (\%) under different demonstration budgets and action-supervision sources.
R and H denote the numbers of robot and human demonstrations per task, respectively.
Do As I Do includes Sharpa-to-xHand adaptation after dynamics-based post-optimization.
Mean is the unweighted average across the three tasks. Conditions shared with
Table~\ref{tab:robot} use the same results.}
\label{tab:robot_budget}
\begin{tabular}{@{}lrrrrrr@{}}
\toprule
Method & Robot R & Human H & Place Vase & Open Box & Insert Toast & Mean \\
\midrule
Robot-only & 20 & 0 & 45.0 & 60.0 & 40.0 & 48.3 \\
Robot-only & 40 & 0 & 60.0 & 65.0 & 55.0 & 60.0 \\
Robot-only & 60 & 0 & 80.0 & 90.0 & 75.0 & 81.7 \\
\midrule
Do As I Do~\citep{paliwal2026dexterous} & 20 & 50 & 50.0 & 45.0 & 40.0 & 45.0 \\
\midrule
Hand2Robot & 20 & 50 & 50.0 & 70.0 & 45.0 & 55.0 \\
Hand2Robot & 20 & 100 & 80.0 & 85.0 & 75.0 & 80.0 \\
Hand2Robot & 0 & 100 & 15.0 & 10.0 & 0.0 & 8.3 \\
\bottomrule
\end{tabular}
\end{table}

%% file: sec/5_conclusion.tex
\section{Conclusion}
\label{sec:conclusion}


In this paper, we presented Hand2Robot, a framework that converts human demonstration
videos into action supervision for dexterous robot policy post-training. The framework
combines global object pose trajectory selection with hand refinement under geometric
and contact constraints, refining hand trajectories and joint configurations before
retargeting them to the robot action space. Experiments on DexYCB and HO3D-v3 show
improved object pose accuracy and hand-object interaction quality. Across three
real-robot dexterous manipulation tasks, adding 100 human demonstrations processed by
Hand2Robot to 20 robot demonstrations per task increases the mean success rate from
48.3\% to 80.0\%, demonstrating the value of reconstructed human demonstrations as a
complement to limited robot data for policy post-training. Future work will extend the
framework to deformable objects and more dynamic contact-rich manipulation.

%% file: sec/6_acknowledgments.tex
\section*{Acknowledgments}

This work was supported by the National Natural Science Foundation of China
(Grant Nos.~62406195 and 62303319), the HPC Platform of ShanghaiTech University,
the Core Facility Platform of Computer Science and Communication of ShanghaiTech University,
the Key Laboratory of Intelligent Perception and Human-Machine Collaboration
(ShanghaiTech University), Ministry of Education, and the Shanghai Engineering
Research Center of Intelligent Vision and Imaging.

%% file: sec/7_appendix.tex

\appendix

\section*{Appendix}

\input{sec/appendix/method_details}
\input{sec/appendix/hyperparameters_selection}
\input{sec/appendix/evaluation_protocal}
\input{sec/appendix/extended_reconstruction_evaluation}

\input{sec/appendix/quality_results}
\clearpage
\input{sec/appendix/real_world_robot_experiment_details}

%% file: sec/appendix/method_details.tex

\section{Method Details}
\label{app:method_details}

\subsection{Prompt for VLM-based Segmentation and Annotation}

Due to its length, we do not include the full prompt inline in the appendix, as doing so would
substantially reduce readability. Instead, we provide the complete prompt template as a separate
text file, \texttt{prompt.txt}, in the supplementary material. The file contains the exact prompt
used for segmenting human demonstration videos into single-interaction task clips and annotating each clip
with task descriptions, directly contacted objects, hand labels, and concise interaction summaries.

\subsection{Interpretation of IoU-Based Pose Weights}

The IoU term measures agreement between the rendered object silhouette and the observed mask.
High IoU indicates close silhouette agreement, although distinct poses can produce similar
projections, particularly for symmetric objects. Conversely, partial visibility and segmentation
errors can reduce IoU even for an accurate pose. We therefore combine this observation-agreement
weight with ScoreNet scores and temporal transition costs when selecting the object trajectory.

\subsection{Observation-to-Contact Hand Refinement}
\label{app:hand_refinement}

This section provides implementation details for the hand refinement module described in the main
paper. Given per-frame MANO initialization from WiLoR~\citep{potamias2025wilor} and a fixed object
trajectory, we optimize the hand trajectory and pose while keeping the object mesh and object poses
fixed.

\textbf{Inputs and Parameterization}

For each video segment, the module takes as input the initial MANO hand parameters, hand and object
masks, camera intrinsics, a reconstructed object mesh $\Omega=(\mathcal{V}_o,\mathcal{F}_o)$, an
object SDF, and the object poses $\mathcal{O}_t=(R_t^o,T_t^o)$. The hand at frame $t$ is represented
by
\begin{equation}
\Theta_t=\{s_t,\tau_t,\omega_t,\theta_t,\beta_t\},
\end{equation}
where $s_t$ is the global hand scale, $\tau_t$ is the hand translation, $\omega_t$ is the global
orientation, $\theta_t$ is the MANO articulation parameter, and $\beta_t$ is the MANO shape
parameter. During hand refinement, we optimize $s_t$ together with the other MANO parameters,
while keeping the metric-scale object mesh from SAM~3D and its pose trajectory fixed.

\textbf{Action Stage Segmentation and Anchor Initialization}


We identify the transition from grasping to interaction using object motion and hand--object
proximity. Object poses are first projected to valid $SE(3)$ transforms and lightly smoothed.
Let $T_t^o$ be the object translation and $d_t^{\mathrm{ho}}$ the minimum hand--object distance.
The interaction score is
\begin{equation}
e_t=
\alpha\,\frac{\|T^o_t-T^o_{t-1}\|_2}{\delta_{\mathrm{trans}}}
+\beta\,\max\left(0,1-\frac{d_t^{\mathrm{ho}}}{\delta_{\mathrm{ho}}}\right),
\end{equation}
where $\alpha,\beta$ are cue weights and $\delta_{\mathrm{trans}},\delta_{\mathrm{ho}}$ are
normalization thresholds. We select the earliest frame satisfying $e_t>\delta_{\mathrm{on}}$
as the interaction onset and anchor $t_a$. Frames before $t_a$ are labeled as grasping,
and frames from $t_a$ onward are labeled as interaction.

Before batch optimization, we optimize the anchor frame with the single-frame optimizer and compute
a segment-level translation offset
$\Delta\tau=\tau_a^\star-\tau_a^0,$
where $\tau_a^0$ and $\tau_a^\star$ are the initial and refined translations at the anchor. This
offset is added to all keyframes in the segment.

\textbf{Batch Objective}

After anchor initialization, keyframes in the same action segment are optimized jointly. The
objective is
\begin{equation}
\mathcal{L}=
\lambda_{\mathrm{2d}}\mathcal{L}_{\mathrm{2d}}
+\lambda_{\mathrm{match}}\mathcal{L}_{\mathrm{match}}
+\lambda_{\mathrm{temp}}\mathcal{L}_{\mathrm{temp}}
+\lambda_{\mathrm{sdf}}\mathcal{L}_{\mathrm{sdf}}
+\lambda_{\mathrm{sm}}\mathcal{L}_{\mathrm{sm}}
+\lambda_{\mathrm{reg}}\mathcal{L}_{\mathrm{reg}} .
\end{equation}
Here $\mathcal{L}_{\mathrm{2d}}$ aligns hand masks, $\mathcal{L}_{\mathrm{match}}$ enforces
per-frame ray contact, $\mathcal{L}_{\mathrm{temp}}$ stabilizes contact over time,
$\mathcal{L}_{\mathrm{sdf}}$ penalizes penetration and encourages surface contact, and
$\mathcal{L}_{\mathrm{sm}}$ and $\mathcal{L}_{\mathrm{reg}}$ regularize temporal smoothness and
deviation from the initialization.

\textbf{Stage-Weighted 2D Mask Alignment}

During grasping, contact is not yet available, so image evidence is the main constraint. We render
the hand mask $\hat{M}_h^t$ and compare it with the observed hand mask $M_h^t$ using a soft IoU
loss:
\begin{equation}
\mathcal{L}_{\mathrm{2d}}=
\frac{1}{B}\sum_{t=1}^{B}
\gamma_t
\left(
1-
\frac{\langle \hat{M}_h^t,M_h^t\rangle+\epsilon}
{\|\hat{M}_h^t\|_1+\|M_h^t\|_1-\langle \hat{M}_h^t,M_h^t\rangle+\epsilon}
\right),
\end{equation}
where $B$ is the number of optimized frames, $\epsilon$ stabilizes the division, and
$\gamma_t=\gamma_g>1$ for grasping frames and $1$ otherwise.

\textbf{Ray-Based Contact Matching.}

During interaction, the hand should be constrained by object geometry. We follow the ray-based
2D-to-3D contact construction from REISOM~\citep{lin2025zero}. Let $\mathcal{Z}$ be the MANO
contact-zone vertex set, and let $v_{t,k}$ denote the 3D position of the $k$-th contact vertex at
frame $t$. A contact candidate is active if its 2D projection falls inside the object mask:
\begin{equation}
\mathcal{C}_t=\{k\in\mathcal{Z}\mid \Pi_K(v_{t,k})\in M_o^t\}.
\end{equation}
For each active candidate, we cast a camera ray through the projected pixel and intersect it with
the posed object mesh $\Omega_t$ to obtain an object-side surface point $o_{t,k}$. The matching loss
is
\begin{equation}
\mathcal{L}_{\mathrm{match}}=
\frac{1}{|\mathcal{I}_{\mathrm{inter}}|}
\sum_{t\in\mathcal{I}_{\mathrm{inter}}}
\frac{1}{\max(|\mathcal{C}_t|,1)}
\sum_{k\in\mathcal{C}_t}\|v_{t,k}-o_{t,k}\|_2 .
\end{equation}
This loss provides a 3D contact target supported jointly by the object mask and the current object
pose.

\textbf{Object-Local Temporal Contact Stabilization.}

Per-frame ray matching may assign the same hand candidate to different object regions over time. We
therefore compare contact points in the object coordinate frame:
\begin{equation}
\bar{o}_{t,k}=(R_t^o)^\top(o_{t,k}-T_t^o),\qquad
\bar{v}_{t,k}=(R_t^o)^\top(v_{t,k}-T_t^o).
\end{equation}
For candidates active in adjacent interaction frames,
$\mathcal{C}_{t,t+1}=\mathcal{C}_t\cap\mathcal{C}_{t+1}$, we define a stop-gradient local anchor
\begin{equation}
a_{t,k}=\frac{1}{2}\left(\operatorname{sg}(\bar{o}_{t,k})+\operatorname{sg}(\bar{o}_{t+1,k})\right),
\end{equation}
and minimize
\begin{equation}
\mathcal{L}_{\mathrm{temp}} =
\sum_{\substack{t:\; t,t+1\in\mathcal{I}_{\mathrm{inter}}}}
\sum_{k\in\mathcal{C}_{t,t+1}}
\left[
\rho(\|\bar{v}_{t,k}-a_{t,k}\|_2)
+\rho(\|\bar{v}_{t+1,k}-a_{t,k}\|_2)
\right],
\end{equation}
where $\operatorname{sg}$ denotes stop-gradient and $\rho(x)=\sqrt{x^2+\epsilon}$ is a smooth
distance penalty with a small constant $\epsilon$. In implementation, we discard non-finite pairs
and pairs with large object-local distances to avoid unstable ray correspondences.

\textbf{SDF Warmup, Smoothness, and Regularization}\par\nopagebreak[4]

SDF contact and penetration losses are enabled only after a warmup stage. This prevents early SDF
gradients from conflicting with image alignment and ray contact when the initial hand is still far
from the object. After warmup, the SDF term removes penetration and improves surface-level contact.
The smoothness term is computed over keyframes and penalizes high-frequency changes in hand pose,
orientation, and translation. We use second-order smoothness by default, which suppresses
acceleration while allowing approximately constant-velocity motion. The regularization term keeps
scale, translation, pose, and shape close to the initial foundation-model prediction.

\subsection{World-to-Robot Coordinate Alignment}

After retargeting the reconstructed human hand motion, the resulting trajectories are still
expressed in the camera coordinate system. We first transform all reconstructed hand and object
trajectories into a shared world coordinate frame using the estimated camera extrinsics. To obtain
the world-to-robot coordinate transformation, we estimate a representative initial object pose
for the human and robot demonstrations, respectively. For the human demonstrations, we collect
the first-frame object pose of each valid sequence in the world frame. The representative
translation is the mean of a Gaussian fitted to these initial positions, while the representative
rotation is computed by averaging on the rotation manifold. For the robot demonstrations, we
estimate object poses using the camera in the robot setup and transform them into the robot
coordinate frame through hand-eye calibration, then compute the corresponding representative
initial object pose. The rigid transformation mapping the human-domain representative pose to
the robot-domain representative pose is used as the world-to-robot alignment matrix. Applying
this transformation expresses the global poses of the reconstructed and retargeted human
trajectories in the same coordinate system as the robot demonstrations.

%% file: sec/appendix/hyperparameters_selection.tex

\section{Hyperparameters Selection}
\label{app:hyperparameters_selection}

\subsection{Hyperparameters for Object Pose Tracking}

In the object pose estimation stage, we set the number of pose candidates to ($N=252$). Following
FoundationPose, these candidates are generated by uniformly sampling object viewpoints and in-plane
rotations, thereby providing broad coverage of the rotation space before pose refinement and
scoring. For the temporal candidate selection objective, we set ($\lambda_\text{trans}=1$) and
($\lambda_\text{rot}=10$). These weights are not task-specific tuned parameters; they are used only
to balance the numerical scales of different terms in the transition cost. We use the same
hyperparameters across all experiments.

\subsection{Hyperparameters for Hand Refinement}
\paragraph{Hand refinement hyperparameters.}
We run hand refinement on an NVIDIA 4090 GPU. Each action segment is optimized with Adam for $400$
steps, using fixed learning rates of $1\times10^{-5}$ for hand scale, $2\times10^{-3}$ for
translation, $1\times10^{-4}$ for global orientation, $2\times10^{-4}$ for articulation, and
$1\times10^{-4}$ for MANO shape. We set $\lambda_{\mathrm{2d}}=10$, $\lambda_{\mathrm{sdf}}=10$,
$\lambda_{\mathrm{reg}}=6$, and $\lambda_{\mathrm{temp}}=8$. The ray contact weight
$\lambda_{\mathrm{match}}$ is $30$ for the first $100$ SDF warmup steps and $20$ afterwards. The SDF
contact base weight is $5$, the grasping-stage mask multiplier is $\gamma_g=2$, and the second-order
smoothness weight $\lambda_{\mathrm{sm}}$ is $0.5$ with stride $1$.

%% file: sec/appendix/evaluation_protocal.tex

\section{Evaluation Protocol}
\label{app:evaluation_protocal}

\subsection{Object Pose Tracking Metrics}

Object poses are estimated using reconstructed object meshes rather than ground-truth CAD models.
For evaluation, we align the predicted poses with the ground truth based on the first frame of each
trajectory to account for the mismatch between canonical coordinate frames.

For all object pose estimation metrics, we first compute the metric value within each trajectory and
then aggregate the resulting per-trajectory values over the evaluation subset. Let $\textbf{t}$ and
$\hat{\textbf{t}}$ denote ground-truth and predicted translation, $\textbf{R}$ and
$\hat{\textbf{R}}$ denote ground-truth and predicted rotation, respectively. Let $\mathcal{M}$
be a set of $m$ sampled object-surface points. Translation error is reported in centimeters and
rotation error in degrees.

\textbf{Translation Error:}
\begin{equation}
    \Vert \hat{\textbf{t}}-\textbf{t}\Vert_2
\end{equation}

\textbf{Rotation Error:}
\begin{equation}
    \arccos\left(
    \frac{\text{tr}(\hat{\textbf{R}}^\top\textbf{R})-1}{2}
    \right)
\end{equation}

The following ADD and ADD-S definitions normalize surface-point distances by the object
diameter. ADD uses corresponding points, whereas ADD-S uses nearest-point matching to account
for indistinguishable surface configurations. Table~\ref{tab:object} reports the area under the
accuracy--threshold curve (AUC) for each distance, rather than the mean distance itself.

\textbf{ADD:}
\begin{equation}
    \frac{1}{m}\sum_{\textbf{x}\in\mathcal{M}}
    \frac
    {\Vert (\hat{\textbf{R}}\textbf{x}+\hat{\textbf{t}})-(\textbf{R}\textbf{x}+\textbf{t}) \Vert_2}
    {\max_{\textbf{x}_1,\textbf{x}_2\in\mathcal{M}}\Vert \textbf{x}_1-\textbf{x}_2\Vert_2}
\end{equation}

\textbf{ADD-S:}
\begin{equation}
    \frac{1}{m}\sum_{\textbf{x}_1\in\mathcal{M}}
    \min_{\textbf{x}_2\in\mathcal{M}}
    \frac
    {\Vert (\hat{\textbf{R}}\textbf{x}_1+\hat{\textbf{t}})-(\textbf{R}\textbf{x}_2+\textbf{t}) \Vert_2}
    {\max_{\textbf{x}_3,\textbf{x}_4\in\mathcal{M}}\Vert \textbf{x}_3-\textbf{x}_4\Vert_2}
\end{equation}

\subsection{Hand Refinement Metrics}
Following prior hand-object motion reconstruction work, we report Mean Per-Joint Position Error
(MPJPE) for hand pose accuracy and Mean Root Position Error (MRPE) for wrist trajectory accuracy
after first-frame alignment. We use Interpenetration Volume (IV) to measure hand-object
interpenetration, reporting both mean and maximum values, and Jerk Magnitude (JM) to measure
temporal smoothness. MPJPE and MRPE are computed on visible-hand frames, while IV and JM are
computed on interaction frames. All metrics are lower-better, with units mm for MPJPE/MRPE, cm$^3$
for IV, and mm/s$^3$ for JM.

Let $\hat{J}_{t,j}$ and $J_{t,j}$ denote the predicted and ground-truth 3D hand joints at frame $t$,
where $j=0,\ldots,J-1$ and $J=21$. We use $\hat{r}_t=\hat{J}_{t,0}$ and $r_t=J_{t,0}$ as the predicted
and ground-truth wrist/root joints. Pose metrics are computed on visible hand frames $\mathcal{V}$,
while penetration and smoothness metrics are computed on interaction frames
$\mathcal{I}_{\mathrm{inter}}$.

We report root-aligned MPJPE to evaluate hand pose accuracy independently of global wrist
translation:
\begin{equation}
\mathrm{MPJPE} =
\frac{1}{|\mathcal{V}|J}
\sum_{t\in\mathcal{V}}\sum_{j=0}^{J-1}
\left\|
(\hat{J}_{t,j}-\hat{r}_t) - (J_{t,j}-r_t)
\right\|_2 .
\end{equation}
For wrist/root trajectory accuracy, we use first-frame-aligned MRPE. Let $t_0=\min\mathcal{V}$ and
$\tilde{r}_t=\hat{r}_t-\hat{r}_{t_0}+r_{t_0}$. Then
\begin{equation}
\mathrm{MRPE} =
\frac{1}{|\mathcal{V}|}
\sum_{t\in\mathcal{V}}
\|\tilde{r}_t-r_t\|_2 .
\end{equation}

For physical plausibility, we report hand-object interpenetration volume. For each interaction
frame, we approximate the intersection volume with a voxel grid $\mathcal{G}_t$ over the overlapping
bounding box of the hand mesh $\hat{\mathcal{H}}_t$ and object mesh $\hat{\mathcal{O}}_t$:
\begin{equation}
\mathrm{IV}_t =
p^3
\sum_{x\in\mathcal{G}_t}
\mathbf{1}[x\in\hat{\mathcal{H}}_t]\,
\mathbf{1}[x\in\hat{\mathcal{O}}_t],
\end{equation}
where $p$ is the voxel side length and $\mathbf{1}[\cdot]$ is the occupancy indicator.
The resulting volume is reported in cm$^3$.

For temporal smoothness, we compute jerk magnitude on the predicted wrist/root trajectory.
In the definition below, $\tau$ denotes physical time in seconds and wrist positions are
expressed in millimeters:
\begin{equation}
\mathrm{JM} =
\frac{1}{|\mathcal{I}_{\mathrm{inter}}|}
\sum_{t\in\mathcal{I}_{\mathrm{inter}}}
\left\|
\left.\frac{d^3 \hat{r}(\tau)}{d\tau^3}\right|_{\tau=\tau_t}
\right\|_2 .
\end{equation}

%% file: sec/appendix/extended_reconstruction_evaluation.tex

\section{Extended Reconstruction Evaluation}
\label{app:extended_reconstruction_evaluation}

\subsection{Object Reconstruction Quality}
\label{app:object_reconstruction_quality}

We evaluate object reconstruction quality on DexYCB and compare our pipeline with
HOLD~\citep{fan2024hold} and MagicHOI~\citep{wang2025magichoi}.
Chamfer distance (CD) measures bidirectional nearest-surface discrepancy, while surface F-scores
combine the precision and recall of matches within a distance threshold.
Table~\ref{tab:object_fidelity} reports CD in cm$^2$ and F@5 and F@10 in percent.
Our pipeline obtains the lowest CD and the highest F-scores among the compared methods
on the evaluated data.

\input{table/object_fidelity}

\begin{table}[htbp]
\caption{We ablate the object grounding module. ``Valid'' denotes the percentage of sequences with a
valid prediction and non-zero overlap with the ground-truth mask. Metrics are reported as
full/valid, where full counts detection failures and zero-overlap predictions as IoU $=0$, and valid
is computed only over valid sequences.}
\label{tab:grounding}
\resizebox{\linewidth}{!}{
\begin{tabular}{ccccccc}
\toprule
VLM        & Detector   & Valid (\%) $\uparrow$ & mIoU $\uparrow$    & Acc@0.25 (\%) $\uparrow$ & Acc@0.50 (\%) $\uparrow$ & Acc@0.75 (\%) $\uparrow$ \\
\midrule
$\times$           &   $\times$         & 20.2                           & 0.08/0.38          & 13.2/65.3               & 7.3/36.1                & 2.5/12.4                \\
    $\times$       & \checkmark & 84.5                           & 0.56/0.66          & 76.0/89.9               & 70.2/83.1               & 42.3/\textbf{50.1}      \\
\checkmark &      $\times$      & 87.0                           & 0.58/0.67          & 83.5/96.0               & 71.1/81.7               & 38.8/44.6               \\
\checkmark & \checkmark & \textbf{95.8}                  & \textbf{0.66}/\textbf{0.69} & \textbf{93.8}/\textbf{97.9}      & \textbf{81.1}/\textbf{84.7}      & \textbf{44.8}/46.8\\
\bottomrule
\end{tabular}
}
\end{table}

\subsection{Object Grounding Ablation}

We additionally evaluate the object grounding module used before object reconstruction and pose
tracking on DexYCB. The full grounding pipeline consists of three steps: a VLM first predicts the
name of the manipulated object, an open-vocabulary detector then localizes the object according to
the predicted text prompt, and SAM3 finally produces the object mask throughout the video using the
detected bounding box as input. To assess the contribution of each component, we compare four
variants: directly prompting SAM3 with ``object in contact with human hand'', using the same generic
prompt to drive the detector before SAM3, using the VLM-predicted object text to prompt SAM3, and
using the complete VLM-detector-SAM3 pipeline.

All metrics are computed at the sequence level. Specifically, we first compute a local IoU score for
each sequence by averaging the per-frame mask IoU within that sequence, and then aggregate the
resulting sequence-level scores over the dataset. We report mean IoU and Acc@$\tau$, where
Acc@$\tau$ denotes the percentage of sequences whose sequence-level IoU is higher than $\tau$. For
each metric, we report two values in the form full/valid. The full setting includes all sequences
and assigns IoU = 0 to detection failures and zero-overlap predictions. The valid setting is
computed only over sequences for which the method produces a valid prediction with non-zero overlap
with the ground-truth mask. We also report Valid, defined as the percentage of such valid sequences
over the full evaluation set.

As shown in Table~\ref{tab:grounding}, directly prompting SAM3 with a generic interaction
description performs poorly, achieving only 20.2\% valid sequences and an overall mIoU of 0.08. This
indicates that mask propagation alone is insufficient when the target object is not explicitly
localized. Introducing either detector-based localization or VLM-based object naming substantially
improves grounding performance: the detector-only variant increases the valid rate to 84.5\%, while
the VLM-only variant reaches 87.0\%. The full pipeline achieves the best overall performance, with
95.8\% valid sequences, an overall mIoU of 0.66, and the highest full-setting Acc@0.25, Acc@0.50,
and Acc@0.75. These gains show that VLM-based semantic object identification and detector-based
spatial localization are complementary: the VLM provides a more precise object-level prompt, while
the detector supplies an explicit spatial prior for SAM3. Although the detector-only variant obtains
a slightly higher valid-setting Acc@0.75, its lower valid rate and weaker full-setting results
suggest that it is less reliable as an end-to-end grounding module. Overall, the complete design
improves both the coverage of successful grounding and the quality of the resulting object masks.

\subsection{Hand Refinement Ablation}
\textbf{Effectiveness of optimization components.}
We conduct an ablation study on DexYCB to inspect the contribution of the main optimization
components. Table~\ref{tab:dex_abl} reports the full objective and variants with individual losses
removed.

The full objective reflects the combined effect of all optimization constraints. Removing the 2D
loss reduces IV-mean from 1.07 to 0.27, but increases MRPE from 38.49 to 113.17 and JM from 2.49
to 9.93. This highlights the role of image alignment in constraining hand motion and shows that
lower penetration alone does not imply more accurate interaction reconstruction. Removing the ray
contact loss weakens the direct hand-object coupling, making the trajectory less constrained by
interaction evidence. Removing the temporal ray contact loss increases JM from 2.49 to 3.19,
indicating that this constraint improves wrist trajectory smoothness. The full objective balances
geometric accuracy, contact quality, and temporal stability.

\begin{table}[!htbp]
 \centering
 \scriptsize
 \setlength{\tabcolsep}{3pt}
\caption{Ablation of the optimization losses on DexYCB. We report MPJPE, MRPE, and physics-based
measures. The first row uses the full objective, while each following row removes one loss term to
assess the role of each constraint.}
 \label{tab:dex_abl}
 \begin{tabular*}{\textwidth}{@{\extracolsep{\fill}}cccccccc@{}}
 \toprule
 $\mathcal{L}_{\mathrm{2d}}$ & $\mathcal{L}_{\mathrm{match}}$ & $\mathcal{L}_{\mathrm{temp}}$ & MPJPE~$\downarrow$ & MRPE~$\downarrow$ & IV-mean~$\downarrow$ & IV-max~$\downarrow$ & JM~$\downarrow$ \\
 \midrule
 $\checkmark$  & $\checkmark$  & $\checkmark$  & \textbf{7.58} & \textbf{38.49} & 1.07 & 3.25 & \textbf{2.49} \\
 $\times$      & $\checkmark$  & $\checkmark$  & 8.38 & 113.17 & \textbf{0.27} & \textbf{1.55} & 9.93 \\
 $\checkmark$  & $\times$      & $\checkmark$  & 7.64 & 48.70 & 0.53 & 2.27 & 2.53 \\
 $\checkmark$  & $\checkmark$  & $\times$      & \textbf{7.58} & 38.62 & 1.07 & 3.12 & 3.19 \\
 \bottomrule
 \end{tabular*}
\end{table}

\textbf{Additional Baseline.}
To further evaluate the effectiveness of our hand reconstruction refinement, we include MagicHOI as
an additional baseline as shown in Table~\ref{tab:magichoi_comparison}. Since our goal is to compare
different hand refinement strategies under the same reconstruction pipeline, we adopt the hand
optimization module from MagicHOI while keeping all other components identical to our method,
including object reconstruction, object pose estimation, and initialization procedures. This setup
isolates the effect of the hand refinement stage and enables a more controlled comparison between
the two optimization strategies. As shown in Table~\ref{tab:magichoi_comparison}, MagicHOI performs
worse than our refinement across all reported metrics and even underperforms the WiLoR
initialization on several metrics.

\begin{table}[!htbp]
  \centering
  \small
  \setlength{\tabcolsep}{5pt}
\caption{We compare additional baseline MagicHOI with WiLoR initialization and our refined hand
trajectories on HO3D-v3. Best values are highlighted in bold.}
  \label{tab:magichoi_comparison}
  \begin{tabular*}{\textwidth}{@{\extracolsep{\fill}}llccccc@{}}
    \toprule
    \textbf{Dataset} & \textbf{Method}
      & MPJPE $\downarrow$
      & MRPE $\downarrow$
      & IV-mean $\downarrow$
      & IV-max $\downarrow$
      & JM $\downarrow$ \\
    \midrule
    \textbf{HO3D-v3}
      & MagicHOI                          & 5.90 & 22.64 & 13.24 & 48.80 & 1.25 \\
      & WiLoR                          & \textbf{4.22} & 24.22 & 20.11 & 64.82 & 0.86 \\
      & Ours                           & 4.40 & \textbf{16.99} & \textbf{3.42} & \textbf{22.40} & \textbf{0.34} \\
    \bottomrule
  \end{tabular*}
\end{table}

%% file: table/object_fidelity.tex
\begin{table}[!htbp]
    \centering
    \caption{Object reconstruction quality on DexYCB. Lower CD and higher F-scores are better.
    Best values are highlighted in bold.}
    \label{tab:object_fidelity}
    \begin{tabular}{lccc}
        \toprule
        Method & CD (cm$^2$) $\downarrow$ & F@5 (\%) $\uparrow$ & F@10 (\%) $\uparrow$ \\
        \midrule
        HOLD~\citep{fan2024hold} & 19.30 & 33.20 & 54.94 \\
        MagicHOI~\citep{wang2025magichoi} & 2.05 & 45.67 & 67.14 \\
        Ours & \textbf{0.57} & \textbf{49.35} & \textbf{87.54} \\
        \bottomrule
    \end{tabular}
\end{table}

%% file: sec/appendix/quality_results.tex

\section{Qualitative Results}
\label{app:quality_results}

\subsection{Hand-Object Reconstruction on HO3D-v3}

\begin{figure}[htbp]
    \centering
    \includegraphics[width=\linewidth]{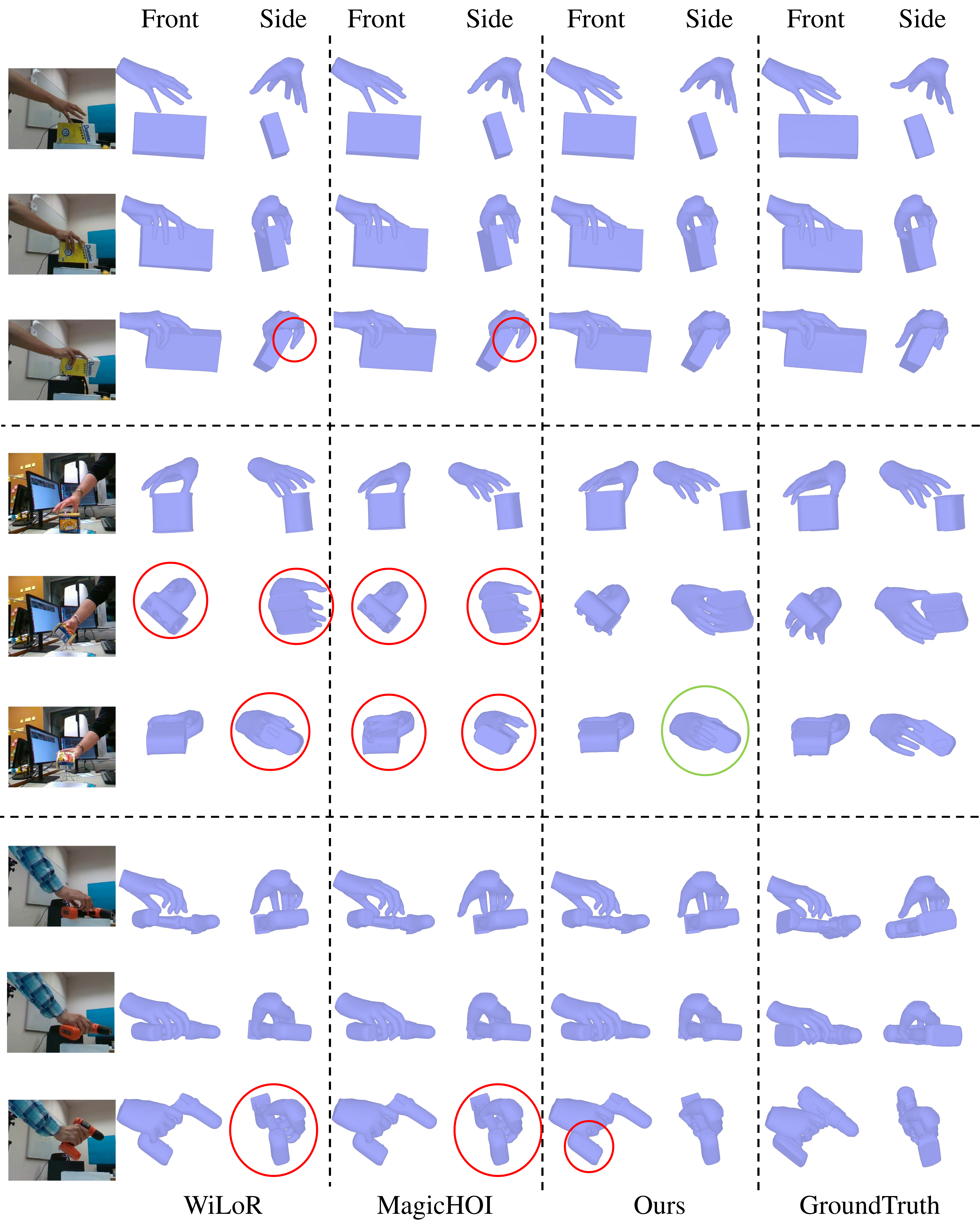}
\caption{Qualitative results of hand-object reconstruction on HO3D-v3. The reconstruction result is
visualized in both the front and side views.}
    \label{fig:qualitative_result_ho3d}
\end{figure}

Figure~\ref{fig:qualitative_result_ho3d} shows qualitative results on HO3D-v3. Our method
consistently improves interaction quality over both the initialization and MagicHOI for objects with
varying geometric complexity, ranging from simple box-shaped objects to challenging drill-shaped
objects. In the drill example, the initialization and MagicHOI do not recover a valid grasp, whereas
our method successfully reconstructs the hand as grasping the drill. Despite slight residual
interpenetration, the reconstructed interaction remains substantially more plausible and stable. The
proposed refinement effectively alleviates hand-object interpenetration and false-contact artifacts,
resulting in more physically consistent reconstructions. As highlighted by the green circles, our
method occasionally produces interaction configurations that appear visually more consistent with
the image observations than the provided annotations.

%% file: sec/appendix/real_world_robot_experiment_details.tex

\section{Real-World Robot Experiment Details}
\label{app:real_world_robot_experiment_details}


\begin{figure}[htbp]
    \centering
    \includegraphics[width=0.4\linewidth]{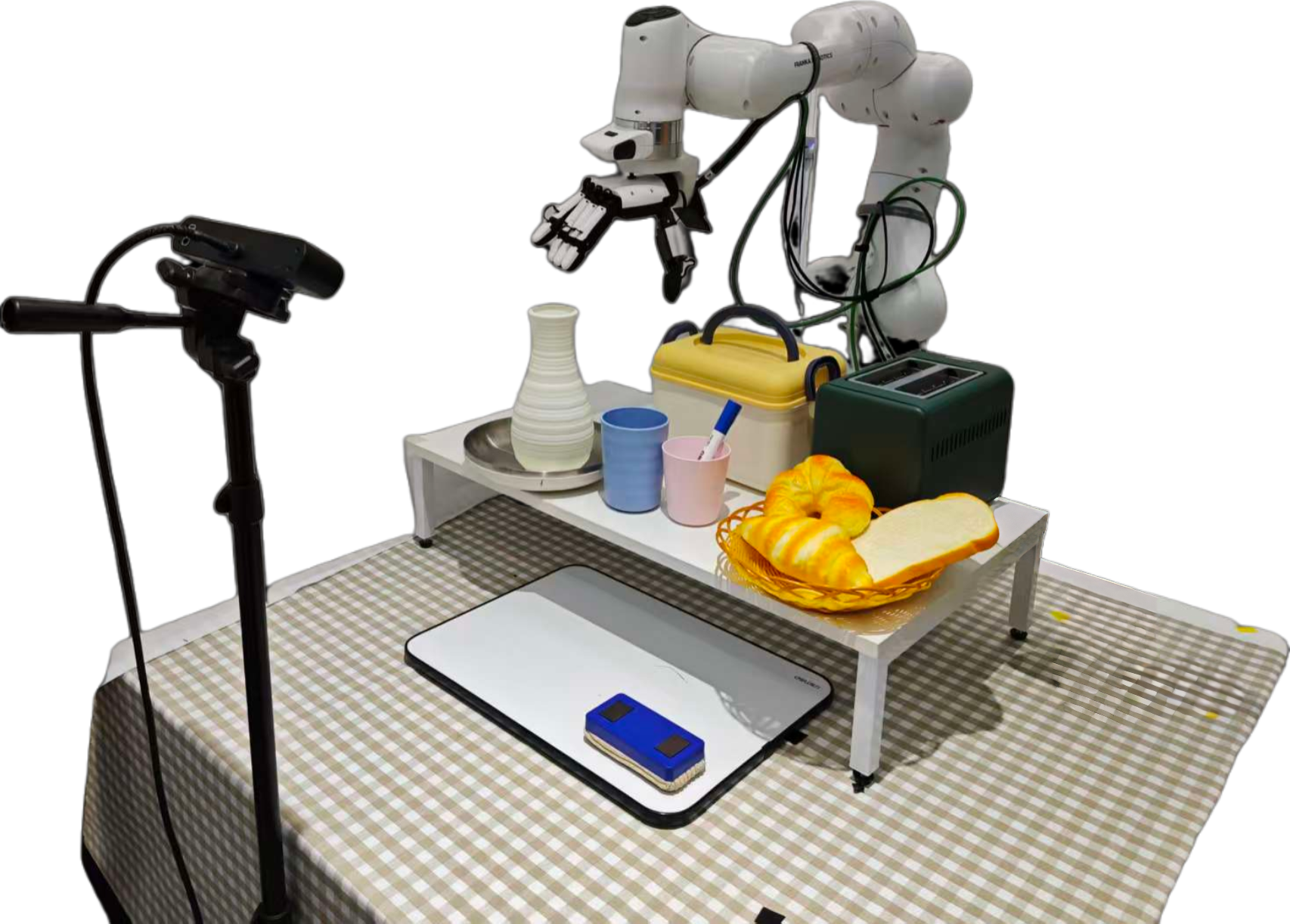}
\caption{Hardware setup of real-world robot experiments.}
    \label{fig:hardware_setup}
\end{figure}

Figure~\ref{fig:hardware_setup} shows the Franka arm and xHand used for the real-robot evaluation.
The current comparisons cover Place Vase, Open Box, and Insert Toast
(Figure~\ref{fig:task}). Human and robot demonstrations are recorded with the same fixed
side-view camera system.

\textbf{Action format.}
The action predicted by the policy consists of two components. The first component represents the
pose of the robot end-effector, including a 3D translation and a 3D rotation parameterized in
axis-angle form. The second component corresponds to the joint angles of the dexterous hand. In our
implementation, we use xHand, which has 12 degrees of freedom. Therefore, the complete action vector
has 18 dimensions. The policy predicts relative action offsets rather than absolute target values.

For WiLoR, MagicHOI, and Hand2Robot, we apply the same xHand retargeting,
coordinate transformation, and relative-action encoding procedures to the recovered
human hand motion before policy post-training.

For Do As I Do~\citep{paliwal2026dexterous}, we retain its dynamics-based
post-optimization and adapt the optimized Sharpa motion to xHand using the same
retargeting method as Hand2Robot.

\textbf{Training data and sampling.}
We post-train $\pi_{0.5}$ for 30,000 steps using the per-task demonstration budgets in
Tables~\ref{tab:robot} and~\ref{tab:robot_budget}.
The action-generation comparison uses the same 20 robot demonstrations and 100 human source
videos per task for WiLoR, MagicHOI, and Hand2Robot.
The budget study evaluates 20, 40, and 60 robot demonstrations, together with mixed-data
conditions using 20 robot demonstrations and either 50 or 100 human demonstrations.
At 20R+50H, Do As I Do and Hand2Robot process the same 50 source human videos per task.
The human-only condition uses 100H.

For all mixed-data conditions, human and robot training samples are drawn at a $1{:}1$ ratio,
regardless of their relative demonstration counts. Robot-only and human-only conditions sample
from their respective data sources.

\textbf{Object placements and evaluation.}
Objects are placed on a tabletop workspace measuring 40~cm $\times$ 70~cm.
In the action-generation comparison in Table~\ref{tab:robot}, human and robot demonstrations
cover the same spatial region. Each trained policy is evaluated in 20 trials per task, with
initial object positions randomly sampled from the training distribution. We compute task
success rate as the number of successful trials divided by 20, and give equal weight to the
three tasks when reporting the mean.

\textbf{Task success.}
Success is judged by task completion: placing the vase on the plate in Place Vase, opening the
box in Open Box, and inserting the toast into the toaster in Insert Toast. The final row of
Figure~\ref{fig:task} illustrates successful completion of the three tasks.